# SubCharacter Chinese-English Neural Machine Translation with Wubi encoding

Wei Zhang[1], Feifei Lin[1], Xiaodong Wang[2], Zhenshuang Liang[2], Zhen Huang[2]
1. Ocean University of China，
2. Global Tone Communication Technology Co., Ltd.(Qingdao)
weizhang@ouc.edu.cn    lff@stu.ouc.edu.cn

**Abstract:** Neural machine translation (NMT) is one of the best methods for understanding the differences in semantic rules between two languages. Especially for Indo-European languages, subword-level models have achieved impressive results. However, when the translation task involves Chinese, semantic granularity remains at the word and character level, so there is still need more fine-grained translation model of Chinese. In this paper, we introduce a simple and effective method for Chinese translation at the sub-character level. Our approach uses the "Wubi method" to translate Chinese into English; byte-pair encoding (BPE) is then applied. Our method for Chinese-English translation eliminates the need for a complicated word segmentation algorithm during preprocessing. Furthermore, our method allows for sub-character-level neural translation based on recurrent neural network (RNN) architecture, without preprocessing. The empirical results show that for Chinese-English translation tasks, our sub-character-level model has a comparable BLEU score to the subword model, despite having a much smaller vocabulary. Additionally, the small vocabulary is highly advantageous for NMT model compression.

## 1 Introduction

Recently, neural machine translation (NMT) (Bahdanau, Cho, and Bengio 2015; Wu et al. 2016; Hassan et al. 2018; Wu et al. 2018; Shen et al. 2016; Luong and Manning 2016) has achieved a very high level of performance. Almost all previous NMT studies used word-, subword- or character-level approaches. Word-level NMT has proven to be remarkably successful (Bahdanau, Cho, and Bengio 2015; Sutskever, Vinyals, and Le 2014). However, due to the fixed vocabulary of languages pairs, the models have difficulty dealing with rare and out-of-vocabulary words. By using byte-pair encoding (BPE) (Gage 1994), subword-level models achieved better performance than word-level models (Sennrich, Haddow, and Birch 2016); to an extent, they benefit from having a small vocabulary, as this helps with the problem of rare words. To achieve open vocabulary NMT, fine-grained models, such as hybrid word-character-level (Luong and Manning 2016; Wu et al. 2016), character-level (Chung, Cho, and Bengio 2016; Ling et al. 2015) and fully character-level models (Lee, Cho, and Hofmann 2017), are typically used and show comparable performance. However, such models are only suitable for Latin language pairs and cannot be directly applied to logographic East Asian languages, such as Chinese.

Chinese translation has proven difficult for machine translation systems (Junczys-Dowmunt, Dwojak, and Hoang 2016). Although they are considered the smallest unit of meaning, Chinese characters can in fact be decomposed into smaller units. For instance, the character 海 (hǎi "sea") is composed of two sub-characters: 氵 (shuǐ "water") and 每 (měi "every"). The character 氵 (shuǐ "water") to some degree reflects the semantics of the character 海 (hǎi "sea"). Similarly, the character 洋 (yáng "ocean") has the same sub-character 氵 (shuǐ "water"). Although it is based on the structure of Chinese characters rather than frequency, this decomposition method is similar to BPE. Similar Chinese characters usually have similar radicals, such that application of BPE to disassembled Chinese characters for language processing tasks is plausible. Glyce (Wu et al. 2019) facilitates sentence encoding; the glyph information of Chinese characters could be used to improve the quality of Chinese translation models, but the current models do not exploit glyph inforSSmation.

The "Wubi method" (Lunde 2009) is a simple method to encode Chinese characters into English characters, while preserving the structural information of the Chinese characters. After processing, the Chinese characters and the derived "Wubi code" have a one-to-one correspondence. (Nikolov et al. 2018) was the first to apply the Wubi method to

Chinese-English translation and achieved promising results in subword- and fully character-level models. They converted raw Chinese characters into Wubi code to eliminate the gap between Chinese characters and alphabetic languages. However, there were some shortcomings associated with their application of the Wubi method. First, when the Wubi method is applied to fully character-level modeling, the model strongly relies on a convolutional neural network (CNN) to reduce the length of the source sentences, making it difficult to apply its architecture directly to newer architectures, such as transformers. In addition, the use of single characters as the basic unit in fully character-level models not only leads to the problem of long sentences (albeit that this can be alleviated through CNN), but also reduces the coherence of Chinese characters. Moreover, when the Wubi code is used in word/subword models instead of Chinese words, the model performance will be affected by word segmentation during pre-processing. Various methods are used for segmenting Chinese words, all of which have their own shortcomings.

In this study, we applied the Wubi method to sub-character Chinese-English recurrent neural network (RNN) translation models (on the Chinese side), which is different from fully character- and subword-level models. We replaced the Chinese characters with the Wubi code, and used BPE to achieve sub-character-level performance without changing the model architecture. The sub-character-level model showed comparable or better performance than the subword-level and fully character-level models. We then explored the influence of BPE on the performance of the sub-character-level model, and demonstrated its superiority over the subword-level model, where the latter is dependent on segmentation of Chinese words. We then showed the advantage of the sub-character-level model in terms of compression. The major contributions of this study are twofold:

(1) We devised a sub-character-level neural translation model based on RNN architecture without the need for explicit segmentation or additional convolutional networks for preprocessing;

(2) We demonstrated the advantage of sub-character-level neural translation with respect to model compression, where the model benefits from a small vocabulary size.

## 2 Related work

(Nguyen, Brooke, and Baldwin 2017) considered the semantics of Japanese characters in their model, employing "sub-character elements" generated by decomposing kanji characters. They found that the decomposition improved the model performance. In a Chinese translation task, (Du and Way 2017) used Pinyin Chinese instead of Chinese characters. However, the method could not adequately deal with Chinese polyphonic characters.

Most recently, (Wu et al. 2019) and (Nikolov et al. 2018) employed the sub-character elements of decomposed Chinese characters in a Chinese-English translation task. (Wu et al. 2019) used CNNs to obtain glyph information from Chinese characters and pictographic information from historical texts; glyph vectors were used to represent the Chinese characters. In Chinese-English translation tasks, use of their method improved the BLEU score by 1.25 (Papineni et al. 2001). However, similar to standard NMT models, the method cannot deal with rare or out-of-vocabulary words. (Nikolov et al. 2018) used the Wubi method to encode Chinese characters. They trained models at the fully character level, as proposed by (Lee, Cho, and Hofmann 2017), and achieved performance comparable to that obtained using raw Chinese characters. However, in fully character-level models, sentences are often too long.

Fine-grained models are helpful for solving the problem of a fixed vocabulary. Thus, we employed the Wubi method and BPE in Chinese-English translation tasks to devise a sub-character-level model without explicit segmentation. The advantages of the sub-character-level model with respect to compression were also examined.

## 3 Background & Method

### 3.1 Neural Machine Translation

We applied the attention-based encoder-decoder model architecture for neural machine translation; an end-to-end neural network was used to transform source sentences into target sentences (Bahdanau, Cho, and Bengio 2015; Sutskever, Vinyals, and Le 2014). Our method can also be applies to other architectures. The NMT system has three components: an encoder, a decoder and an attention mechanism.

**Encoder** - Given a source sentence $(x_1, x_2, ..., x_n)$ of length n, the encoder bidirectionally encodes the hidden states of each position i to $\vec{h}_i$ and $\overleftarrow{h}_i$ with a gated recurrent unit (GRU) or long short-term memory (LSTM) RNN, via Equation 1:

$$\vec{h}_i = frnn(E(x_i), \vec{h}_{i-1})$$
$$\overleftarrow{h}_i = frnn(E(x_i), \overleftarrow{h}_{i-1})$$
(1)

where $frnn$ is the RNN function and $E_x$ is the embedding matrix. The hidden states are then concatenated into a bidirectional representation $h_i$.

**Decoder** - Based on the source sentence representation and the embedding lookup table, the decoder predicts the target sentence $(y_1, y_2, ..., y_m)$ through the RNN. When predicting the target word $y_j$, except for the current hidden state $s_j$ and the previously predicted word $y_{j-1}$, the decoder always considers the attention mechanism.

**Attention Mechanism** - Proposed by (Bahdanau, Cho, and Bengio 2015) and extended by (Luong, Pham, and Manning 2015), the attention mechanism attends to different source tokens when the decoder generates target tokens at each time step. To summarize the source attentional words, it uses a context vector, which is calculated through the hidden states $(h_1, h_2, ..., h_n)$ in the encoder shown in Equation 2:

$$c_t = \alpha_t h$$
(2)

where $\alpha_t$ is the attention vector computed by Equation 3:

$$\alpha_t = softmax(a_t)$$
(3)

where $a_t$ is the attention score computed according to the encoder's hidden states $h$ and the previous decoder state $s_{t-1}$.

## 3.2 Byte-pair Encoding
BPE is an efficient data compression algorithm that is widely used in NMT for word segmentation, to achieve subword-level models (Sennrich, Haddow, and Birch 2016). First, BPE iteratively replaces the most frequent characters with a new symbol which is the merging of several most frequent characters until the required vocabulary size is reached. Since the final tokens in the vocabulary are the most frequent word fragments, NMT models based on BPE not only show vocabulary compression, but also process rare words reasonably well. For two similar languages, the joint BPE method can be used to generate the vocabularies for both the source and target language.

As the vocabulary size decreases, despite the longer sentence lengths, subword-level models show better performance. We employ BPE in sub-character-level models to further improve their ability to deal with rare words.

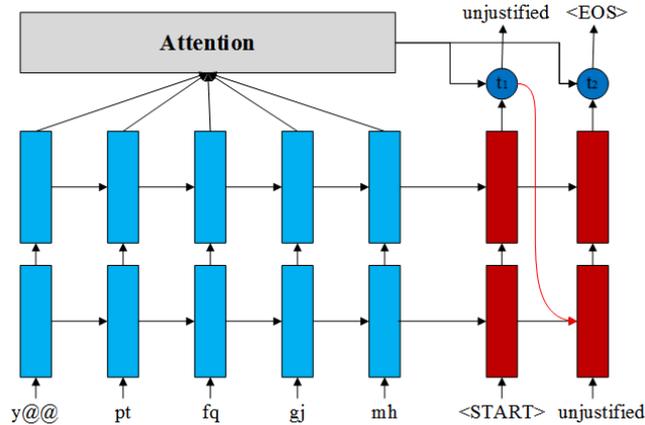

Figure 1: Sub-character level NMT – an example for translating "毫无理由" into "unjustified". "y@@ pt fq gj mh" is the wubi sequence of "毫无理由". Sub-character level NMT uses natural boundary to split Chinese characters. <START> and <EOS> mark sequence boundaries.

## 3.3 Sub-character level translation
A Chinese character is the smallest unit of meaning in Chinese, and most previous studies of Chinese translation tasks were performed at the word, subword or character level. Using the Wubi method and BPE, we decompose Chinese characters into sub-characters for sub-character-level NMT (see Fig.1). For example, 'ypt' is the Wubi code for the Chinese character '毫'; when BPE is used, 'ypt' is decomposed into two parts, 'y@@' and 'pt'. Unlike the word-level model, the sub-character-level model exploits the natural boundaries of each Chinese character rather than relying on explicit word segmentation. Chinese word segmentation is more impactful with respect to translating Chinese, but is usually a very complicated procedure. Different methods of word segmentation produce different vocabularies, which will affect the characteristics of the corpus to be learned.

For example, it is difficult to select the correct word from a sentence that contains multiple instances of the same Chinese character. Without appropriate word segmentation, the translation quality will be reduced. Using the natural boundary of Chinese characters for segmentation eliminates the uncertainty associated with Chinese word segmentation. Although the BPE method used in our model also has some uncertainty, it is different from that of Chinese word segmentation. BPE further processes the corpus to achieve open vocabulary NMT.

### 3.3 Wubi Method

The Wubi method is a Chinese character input method that encodes Chinese characters into 25 English letters (a–y), by reference to the radicals of Chinese characters. Each Chinese character is losslessly encoded into a sequence of one to five English letters (typically four letters). The Chinese characters are decomposed based on their structure, which to some degree reflects their semantics.

Wubi divides the 25 aforementioned English letters into five groups, which represent the Chinese characters of 一 (horizontal, 'heng'), 丨 ( vertical, 'shu'), 丿 (skim, 'pie'), 丶 (restrain, 'na'), and 乙 (fracture, 'zhe'). Each group contains a number of radicals. Chinese characters are distinguished according to their roots, and are then encoded into Wubi characters. For example, '照' is encoded as 'jvko', in which 'j' is '日', 'v' is '刀', 'k' is '口' and 'o' is '灬'. A small number of Chinese characters share the same Wubi code, but unique codes are ensured by adding numbers to the end of such characters. In our sub-character-level experiment, we converted each Chinese character into Wubi code. Table 1 gives an example wherein Chinese1 is the raw sentence, Chinese2 is the sentence split by natural boundaries, the third row is the sentence encoded by the Wubi method, and the last row is the English translation.

Table 1: An example of Wubi method.

| Chinese1 | 这种说法毫无根据。 |
|---|---|
| Chinese2 | 这 种 说 法 毫 无 根 据 。 |
| Wubi | p tkh yu if ypt fq sve rndg . |
| English | This claim is groundless . |

# 4 Experiment

All experiments involved Chinese-English translation tasks. The sub-character- and subword-level models were trained with Nematus[1] (Sennrich et al. 2017), which has an attention-based encoder-decoder architecture, and the fully character-level model was trained with char2char[2] (Lee, Cho, and Hofmann 2017). We trained the subword- and fully character-level models only for comparison. We used the multi-bleu.perl Moses script[3] to calculate BLEU scores for final model evaluation, and we use word-level BLEU scores for all models.

### 4.1 Data and Preprocessing

The parallel corpora used for Chinese-English translation were from the United Nations Parallel Corpus (Ziemski, Junczys-Dowmunt, and Pouliquen 2016). Our objectives were to verify the availability of the sub-character-level model and determine the advantages of fine-grained models. We extracted 500,000 sentences as a training set, 2,000 sentences as a development set, and 2,000 sentences as a test set. Then, we omitted sentences that contained foreign language characters. Following this, we removed all sentences that contained numbers. Detailed information regarding the training sets are shown in Table 2. For the subword-level model that required Chinese word segmentation, we used the monolingual corpus of the China Workshop of Machine Translation (CWMT) to train language models to segment Chinese words.

Table 2: Statistics information of training data, tokens is the word and character countes. We use Jieba for Chinese word segmentation for comparison models.

|  | Chinese | | English | |
|---|---|---|---|---|
|  | word | character | word | subword |
| Sents | 500,000 | | | |
| Tokens | 9.6M | 17.1M | 11.3M | 11.6M |

For the sub-character-level models, Chinese characters were encoded using the Wubi method[4] (to generate Wubi code) and then processed like English words. We also converted Chinese punctuation into English punctuation (e.g., '。' to '.'), similar to (Nikolov et al. 2018). For the word-level models, we used two methods for word segmentation: Jieba[5], and a word segmentation model based on a language model. We trained different language models for word segmentation, mainly to determine the influence of word segmentation on model quality. Note that Chinese word segmentation was only applied as a pre-processing step in the word-level NMTs; our sub-character-level model does not require word segmentation. For the English corpus, we applied tokenization and truecasing using Moses scripts.

---

[1] https://github.com/EdinburghNLP/nematus/tree/theano
[2] https://github.com/nyu-dl/dl4mt-c2c
[3] https://github.com/moses-smt/mosesdecoder
[4] https://github.com/arcsecw/wubi

For training, the word-level model had a vocabulary size of 30,000 words; BPE operations were applied to the English corpus 20,000 times to build the vocabulary.

### 4.2 Task and Models

We conducted two types of translations: Chinese-English and Wubi-English. The Chinese-English translation model considered both the word and subword levels, while the Wubi-English translation model was trained at the sub-character-level and fully character-level. Different numbers of BPE operations (0, 500, 1,000, 2,000, 3,000 or 4,000)[6] were applied to the Chinese corpus during preprocessing and training of the sub-character-level models. The experiments included two tasks.

**Task 1** In this task, the objective was to assess the performance of the Wubi method when applied to the sub-character-level models, and to determine the relationship between Chinese vocabulary size and model quality. We used different numbers of BPE operations to train the models, as detailed above. Then, we compared the sub-character- and subword-level models, the latter of which use various word segmentation methods, to demonstrate the advantage of sub-character-level models (which do not use word segmentation). The sub-character- and fully character-level models were also compared.

**Task 2** In this task, we aimed to evaluate the advantages of fine-grained NMT models. The Wubi method can significantly reduce the size of the Chinese vocabulary used in the model, and we aimed to determine whether this was helpful as a model compression. We reduced the sizes of the embedding and hidden layers to train the models, and noted changes in the BLEU scores and size of the models.

### 4.3 Training Details

We applied the same model-training process throughout so that the parameters used for assessing model quality were as consistent as possible.

During training, the word embedding and hidden layer sizes were set to 512 and 1,024, respectively. A checkpoint was created after every 30,000 iterations. For optimization, Adadelta was used (Zeiler 2012). The initial learning rate was set at 0.0001. We used shuffle for each epoch in training. The batch sizes of the training set and validation set were set to 64 and 32, respectively. The dropout rates for the word embedding, hidden, source and target layers were set to 0.2, 0.2, 0.1 and 0.1, respectively. All neural networks had two layers. For early-stop, we set patience to 30. Layer normalization was used to improve the model training speed. For the fully character-level model, we used the same parameters employed by (Nikolov et al. 2018). All models were trained on a computer with an NV GTX 1080 GPU.

## 5 Analysis

### 5.1 Evaluation with BLEU Scores

Table 3 shows the BLEU scores of all models for the development set and the test set. On the ZH-EN translation task, the performance of the sub-character-level models was generally better than that of the character models. When there were 2,000 BPE operations, the sub-character-level models yielded the best BLEU scores for both the development set and the test set (41.40 and 36.46, respectively). Interestingly, the score for the best sub-character-level model was 0.33 lower than that for the subword-level model for the development set, and 0.14 higher for the test set. The subword-level model may benefit from its large vocabulary. The sub-character-level models performed similarly on the EN-ZH and ZH-EN translation tasks. For the subword-level model, the BLEU score was 0.41 higher for the development versus test set, while the best sub-character-level model had a BLEU score that was 0.2 higher for the test versus development set. The BLEU scores were very similar among the different models. We suspect that this is due to the learning ability of the RNN architecture itself. We found that when compared with the character model (wubi:0), almost all of the sub-character-level models had higher BLEU scores, which not only verifies the effectiveness of BPE, but also demonstrates the effectiveness of the sub-character-level models themselves. The sub-character-level models improved upon the character model, achieving a level of performance comparable to that of the subword- and fully character-level models.

---
[5] https://github.com/fxsjy/jieba
[6] The numbers in bracket is the BPE operation for Chinese corpus, and 0 means no BPE.

Table 3: The BLEU scores of all models. (ZH-EN : Chinese ⟶ English; EN-ZH : English ⟶ Chinese)

| Model | | ZH-EN | | EN-ZH | |
|---|---|---|---|---|---|
| | | dev | test | dev | test |
| word | | 41.16 | 36.11 | - | - |
| subword | | 41.73 | 36.53 | 37.11 | 35.58 |
| wubi | 0 | 40.98 | 36.21 | 36.76 | 35.33 |
| | 4000 | 41.33 | 36.48 | **36.70** | 35.77 |
| | 3000 | 41.27 | 36.38 | 36.63 | **35.78** |
| | 2000 | **41.40** | **36.67** | 36.66 | 35.56 |
| | 1000 | 40.93 | 35.97 | 36.42 | 35.18 |
| | 500 | 40.87 | 35.76 | 36.36 | 35.23 |

The performance of the sub-character-level models relies on BPE on the Chinese side. Fig. 2 shows the BLEU scores (for the test set) of the sub-character-level models with different numbers of BPE operations. For comparison purposes, Fig. 2 also contains the results of the character model (wubi:0), which does not use the BPE method. From Fig. 2, we can see that the performance of the sub-character-level model was not identical between the EN-ZH and ZH-EN tasks. On the ZH-EN task, 2,000 BPE operations was optimal. On the EN-ZH task, 3,000 BPE operations was optimal. With only 500 BPEs, the BLEU scores for the test set dropped by 0.91 and 0.55, respectively, for the ZH-EN and EN-ZH tasks. We can conclude that BPE corresponding to approximately 60% of the raw vocabulary size appears optimal for improving the quality of sub-character-level models.

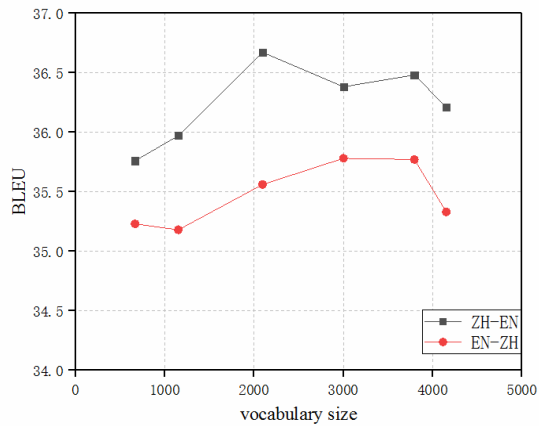

Figure 2: BLEU scores on the test set of different sub-character level models. (The last column is the result of character model that does not use BPE method.)

Table 4 shows the highest BLEU scores for the test set obtained by the sub-character and fully character-level (Nikolov et al. 2018) models. The BLEU score for the sub-character-level model is higher than that for the fully character-level model, indicating that the sub-character method is superior.

### 5.2 Compare with word segmentation
One advantage of the sub-character-level model is that it does not require word segmentation. To determine the impact of word segmentation on model quality, we trained the models using different word segmentation methods. We trained two language models for Chinese word segmentation, differing in corpus size (i.e., in "quality"; 1 vs. 2 million sentences). We also trained the character-level model which uses natural boundaries for word segmentation. Jieba was used for model comparison. We trained NMT models on the ZH-EN translation task and present the results in Table 5.

Table 4: The best BLEU scores on the test set of sub-character and fully-character models. (Note that the BLEU scores of all models are in word-level)

| Model | ZH-EN | EN-ZH |
|---|---|---|
|  | test | test |
| fully-character | 24.57 | 23.67 |
| sub-character | 36.67 | 35.78 |

Table 5: The BLEU scores of subword model with different word segmentation. (character means use natural boundary for word segmentation)

| scheme | words/sent | BLEU(ZH-EN) | |
|---|---|---|---|
|  |  | dev | test |
| lm01 | 19.89 | 36.74 | 36.15 |
| lm02 | 19.72 | 36.92 | 36.44 |
| Jieba | 19.32 | 41.73 | 36.53 |
| character | 34.34 | 41.40 | 36.67 |

From the results shown in Table 5, it is apparent that different word segmentation methods produce sentences of different lengths, and the higher the quality of the word segmentation, the smaller the average number of words. The average sentence length obtained by the word segmentation language model trained with the 1 million-sentence corpus was 19.89 words, and the BLEU score was 36.15. When the amount of training data for the language model was doubled, the average sentence length decreased by 0.17, and the BLEU score for the test set increased by 0.29. The results showed that the better the quality of the language model, the better the quality of word segmentation, and the better the quality of the NMT model. Compared with the character model, the average length of the corpus was up to 34.34; however, the BLEU score was higher. This suggests that sentence length does not adversely affect model learning, although the optimal Chinese word segmentation scheme is difficult to determine. The sub-character/character model based on natural boundaries is simple and effective, without the uncertainty associated with the word segmentation performed for word/subword models.

### 5.3 Evaluation

It is intuitive to expect that models with smaller vocabularies could be trained with fewer network parameters, while still achieving good performance. In this section, we discuss the impact of reducing the size of the word embedding and hidden layers on BLEU scores. The vocabulary sizes (both Chinese and English) of the word- and subword-level models were kept consistent with those of the sub-character-level models.

Table 6: The information of subword- and sub-character-level models (Word embedding size: 256 and hidden size: 512)

| Level | | Embedding | BLEU |
|---|---|---|---|
| ZH:EN | subword | 682K:20.1M | 30.88 |
|  | wubi(500) | 682K:20.1M | 34.21 |
| EN:ZH | subword | 20.1M:682K | 18.34 |
|  | wubi(500) | 20.1M:682K | 34.11 |

Table 6 shows the results. When the number of network parameters was halved, the sub-character-level model performed better than the subword-level model. On the ZH-EN translation task, the BLEU scores for the sub-character- and subword-level models were 34.21 and 30.88, while on the EN-ZH task the BLEU scores were 34.11 and 18.34, respectively. It is clear that the performance of the subword-level model was strongly influenced by the reduction in network parameters and vocabulary size, whereas the sub-character-level model showed high robustness. The ability of the sub-character-level model to retain a high level of performance with fewer parameters is largely attributable to its small vocabulary. Compressing the vocabulary is the only way to achieve a smaller model without changing the model architecture, and can be applied to many models due to its general nature; the sub-character-level model shows proficiency in compressing the vocabulary size.

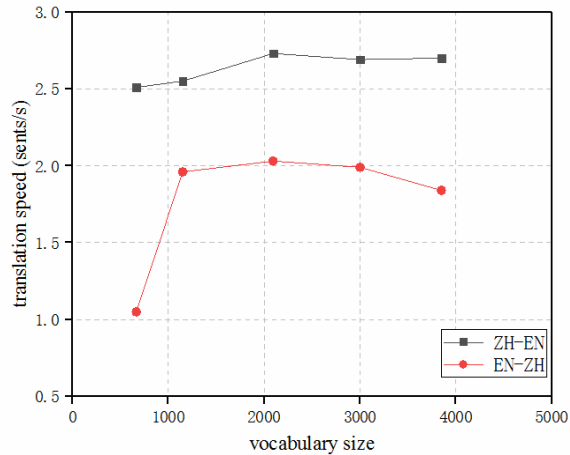

Figure 3: Translation speed versus model vocabulary size.

## 5.4 Speed

Sub-character-level models have higher computational costs than subword- and word-level models, because the sentences that they address are longer; however, the cost is still lower than that of the fully character-level model. Fig. 3 plots the translation speed for sub-character-level models differing in vocabulary size. The translation speed was strongly related to the quality of the models (c.f. Fig. 2 and 3). On the ZH-EN translation task, the highest-quality sub-character-level model, with a vocabulary size of 2,093, achieved the highest translation speed, of 2.73 sentences/second; the subword-level model was slower by 0.4 sentences/second, while the word-level model was faster by 0.09 sentences/second, and all much faster than sub-character-level model (the speeds for the word- and subword-level models are not shown in Fig. 3). On the EN-ZH translation task, with 500 BPE operations, the translation speed was almost halved due to the longer sentence length. The translation speed changed little between 1,000 and 4,000 BPE operations. On both tasks, the average length of the sentences for the sub-character-level models was approximately twice that for the subword- and word-level models. Despite the longer sentences, the translation speed of the sub-character-level models was not significantly slower.

For comparison, we tested the translation speed of the fully character-level model. The speed was 1.18 sentences/second for the ZH-EN translation task and 1.36 sentences/second for the EN-ZH translation task, which was significantly lower than that for the subword- and sub-character-level models. Although the fully character-level model can address the longer sentences via CNN, the decode is still need to be done character by character, which leads to a very slow speed. Therefore, the sub-character-level model has significant advantages in terms of decoding speed.

## 5.5 Sample Translations

Table 7 shows three sample translations from the test dataset generated by different models on the ZH-EN translation task. Here, we focus on the translation generated by the sub-character-level model, and compare it with those of the subword- and word-level models. In the first example, the word- and sub-character-level models translated '是' into 'was', which represents a tense error. The output of the subword-level model fully matches the ground truth.

In the second example, the sub-character-level model showed impressive fluency. It uses the phrase 'in respect of' to connect the two parts of the sentence, which differs from the ground truth text but is superior to the output of the subword- and word-level models. Interestingly, the output for the sub-character- and subword-level models showed similar sentence structure; however, the structure of the word-level model best matches the ground truth.

Lastly, we compared the performance of various models on long sentences, which is more challenging. The sub-character- and word-level models successfully translated the word '安全', which was translated into 'safety' by the subword-level model. The sub-character-level model failed to accurately translate the word '参与'; it was translated as 'participation', which is different from the reference word 'involvement'. In addition, the word-level model failed to translate the word '全体'. The sub-character-level model was the only one model to successfully translate '应对' into 'address'. Meanwhile, the word-level model successfully translated 'impacts'; it was translated into 'effects' by the other models, which has a very similar meaning. We were surprised to find that the sub-character-level model translated the word '海基' to 'ocean-based', whereas it was translated into 'UNK' by the word-level model. This shows the superiority of the sub-character-level model for translating unseen words. The word '海基' was decomposed into '海' and '基', which were translated into "ocean" and "based", respectively, by the sub-character-level model. Regarding the translation results of the sub-character-level model for long sentences, although our sub-character-level model made some mistakes, its performance was comparable to that of the subword- and word-level models.

Table 7: Sample translations from test dataset, for each example, we show the source and target translation and compare the translations of subword-level, word-level and sub-character-level NMT systems. The sub-character-level model's BPE operation is 2,000.

| | | |
|---|---|---|
| 1 | SRC | 教育是创造平等的关键。 |
| | TRG | Education is key in creating equality. |
| | subword | Education is the key to creating equality. |
| | word | Education was the key to creating equality. |
| | wubi | Education was the key to creating equality. |
| 2 | SRC | 迄今为止，对该报告中提出的建议似乎尚无任何后续司法行动。 |
| | TRG | To date, there appeared to have been no judicial follow-up in response to the recommendations of the report. |
| | subword | To date, no follow-up judicial action appears to be found in the recommendations made in the report. |
| | word | To date, there appear to be no follow-up judicial action on the recommendations made in this report. |
| | wubi | To date, no follow-up judicial action appeared to have been made in respect of the recommendations made in the report. |
| 3 | SRC | 他指出，海洋在未来全球食品安全方面的机遇取决于全体利益攸关方应对累积影响的政治意愿和参与，这要求重建并以可持续方式管理渔业，减轻陆基和海基活动所造成的影响，对海洋使用进行综合全面的管理。 |
| | TRG | He noted that opportunities for oceans in future global food security depended on political will and the involvement of all stakeholders to address cumulative impacts, which required rebuilding and sustainably managing fisheries, mitigating impacts from land-based and sea-based activities and managing ocean uses in an integrated holistic manner. |
| | subword | He noted that the opportunities for oceans in future global food safety depended on the political will and involvement of all stakeholders to respond to the cumulative impact, which required the reconstruction and sustainable management of fisheries, the reduction of the impact of land-based and sea-based activities and the comprehensive and comprehensive management of the use of oceans. |
| | word | He noted that the opportunities for oceans for future global food security depended on the political will and involvement of stakeholders in responding to cumulative effects, which required the reconstruction and sustainable management of fisheries, the mitigation of land-based and UNK activities, and the integrated and comprehensive management of ocean use. |
| | wubi | He noted that the opportunities for oceans in future global food security depended on the political will and participation of all stakeholders to address the cumulative effects, which required the reconstruction and sustainable management of fisheries, mitigating the impacts of land-based and ocean-based activities and integrated ocean management. |

# 6 Conclusion

In this paper, we applied the Wubi method to sub-character-level models with RNN architecture for Chinese-English translation tasks, and evaluated the performance of the new models in terms of BLEU scores. We compared sub-character-level models with subword- and fully character-level models in terms of both BLEU scores and translation speed. In addition, we explored the effect of a small vocabulary on model performance. We tested sub-character-level models with different numbers of BPE operations; the word embedding and hidden layer sizes were reduced, and the resulting BLEU scores were compared. Our experimental results showed that:

1. The sub-character-level models, realized by the decomposition of Chinese characters via the Wubi method, are simple and effective, with comparable performance to subword- and fully character-level models. The sub-character-level models accomplished Chinese translation without explicit segmentation and used a relative smaller vocabulary, and were close to achieving open vocabulary.

2. Sub-character-level models have the major advantage that model compression does not seriously degrade translation quality, so model size can be reduced without changing the model architecture.